\documentclass[Crown,sagev,times]{sagej}

\usepackage{moreverb,url}
\usepackage{longtable}

\usepackage[colorlinks,bookmarksopen,bookmarksnumbered,citecolor=red,urlcolor=red]{hyperref}

\newcommand\BibTeX{{\rmfamily B\kern-.05em \textsc{i\kern-.025em b}\kern-.08em
T\kern-.1667em\lower.7ex\hbox{E}\kern-.125emX}}

\begin{document}

\runninghead{Ha and Tang}

\title{Collective Intelligence for Deep Learning: A Survey of Recent Developments}

\author{David Ha\affilnum{1} and Yujin Tang\affilnum{1}}

\affiliation{\affilnum{1}\textit{Google Brain, Tokyo, Japan.\\
Both authors contributed equally to this work.\\
Email: hadavid@google.com, yujintang@google.com}
}



\begin{abstract}
In the past decade, we have witnessed the rise of deep learning to dominate the field of artificial intelligence. Advances in artificial neural networks alongside corresponding advances in hardware accelerators with large memory capacity, together with the availability of large datasets enabled practitioners to train and deploy sophisticated neural network models that achieve state-of-the-art performance on tasks across several fields spanning computer vision, natural language processing, and reinforcement learning. However, as these neural networks become bigger, more complex, and more widely used, fundamental problems with current deep learning models become more apparent. State-of-the-art deep learning models are known to suffer from issues that range from poor robustness, inability to adapt to novel task settings, to requiring rigid and inflexible configuration assumptions. Collective behavior, commonly observed in nature, tends to produce systems that are robust, adaptable, and have less rigid assumptions about the environment configuration. Collective intelligence, as a field, studies the group intelligence that emerges from the interactions of many individuals. Within this field, ideas such as self-organization, emergent behavior, swarm optimization, and cellular automata were developed to model and explain complex systems. It is therefore natural to see these ideas incorporated into newer deep learning methods. In this review, we will provide a historical context of neural network research's involvement with complex systems, and highlight several active areas in modern deep learning research that incorporate the principles of collective intelligence to advance its current capabilities. We hope this review can serve as a bridge between the complex systems and deep learning communities.
\end{abstract}


\keywords{Deep Learning, Reinforcement Learning, Cellular Automata, Self-Organization, Complex Systems}

\maketitle

\section{Introduction}



Deep learning (DL) is a class of machine learning methods that uses multi-layer (``deep'') neural networks for representation learning. While artificial neural networks, trained with the backpropagation algorithm, first appeared in the 1980s~\cite{schmidhuber2014invented}, deep neural networks did not receive widespread attention until 2012 when a deep artificial neural network solution trained on GPUs~\cite{krizhevsky2012imagenet} won an annual image recognition competition~\cite{deng2009imagenet} by a significant margin over the non-DL runner up methods. This success demonstrated that DL, when combined with fast hardware-accelerated implementations and the availability of large datasets, is capable of achieving exceptionally better results in non-trivial tasks than conventional methods. Practitioners soon quickly incorporated DL to address the long-standing problems in several other fields. In computer vision (CV), deep learning models are used in image recognition~\cite{simonyan2014very,he2016deep,radford2021learning} and image generation~\cite{wang2021generative,jabbar2021survey}. In natural language processing (NLP), deep language models can generate text~\cite{radford2018improving,radford2019language,brown2020language} and perform machine translation~\cite{stahlberg2020neural}. Deep learning has also been incorporated into reinforcement learning (RL) to tackle vision-based computer games such as Doom~\cite{ha2018worldmodels} and Atari~\cite{mnih2015human}, and play games with large search spaces such as Go~\cite{silver2016mastering} and Starcraft~\cite{vinyals2019grandmaster}. Deep learning models are also deployed for mobile applications like speech recognition~\cite{alam2020survey} and speech synthesis~\cite{tan2021survey}, demonstrating their wide applicability.

\begin{figure}[ht]
\vskip -0.05in
\begin{center}
\centerline{\includegraphics[width=1.0\columnwidth]{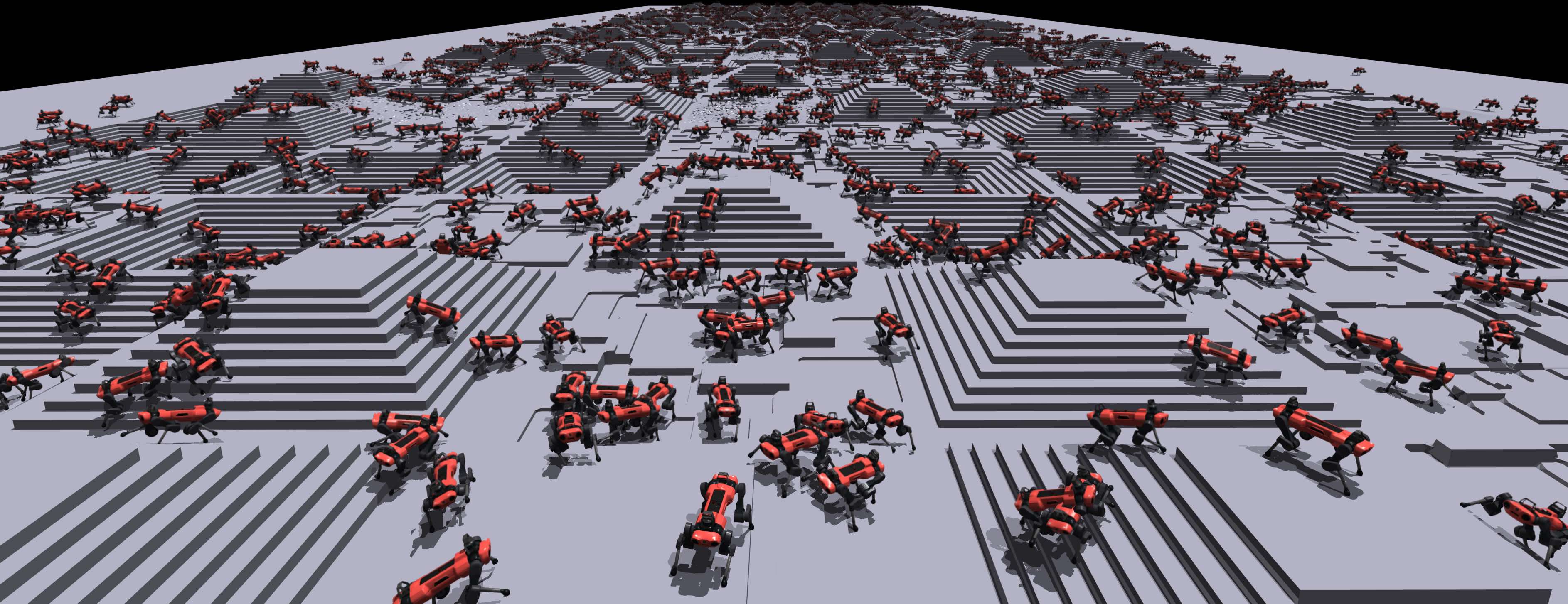}}
\vskip -0.00in
\caption{Recent advances in GPU hardware enables realistic 3D simulation of thousands of robot models~\cite{heiden2021neuralsim}, such as the one shown in this figure from Rudin et al.~\cite{rudin2021learning}. Such advances opens the door for large scale 3D simulation of artificial agents that can interact with each other and collectively develop intelligent behavior.}
\label{fig:rudin2021learning}
\end{center}
\vskip -0.35in
\end{figure}

However, DL is not an elixir without side effects. While we are witnessing many successes and a growing adoption of deep neural networks, fundamental problems with DL are also revealing themselves more and more clearly as our models and training algorithms become bigger and more complex. DL models are not robust in some cases. For example, it is now known that by simply modifying several pixels on the screen of a video game (the modification is not even noticeable to humans), the agent trained with unmodified screens that originally surpassed human performance could fail~\cite{qu2020minimalistic}. Also, CV models trained without special treatment may fail to recognize rotated or similarly transformed examples, in other words, our current model and training methods do not lend themselves to generalization to novel task settings. Last but not least, most DL models do not adapt to changes. They make assumptions about input and expect rigid configurations and stationarity of the environment, what statisticians think of as the data generating process. For instance, they may expect a fixed number of inputs, in a determined ordered. We cannot expect agents to capably act beyond their skills learned during training, but once these rigid configurations are violated, the models do not perform well unless we retrain them or manually process the inputs to be consistent with the expectations of their initial training configurations.

\begin{figure}[ht]
\vskip -0.15in
\begin{center}
\centerline{\includegraphics[width=0.75\columnwidth]{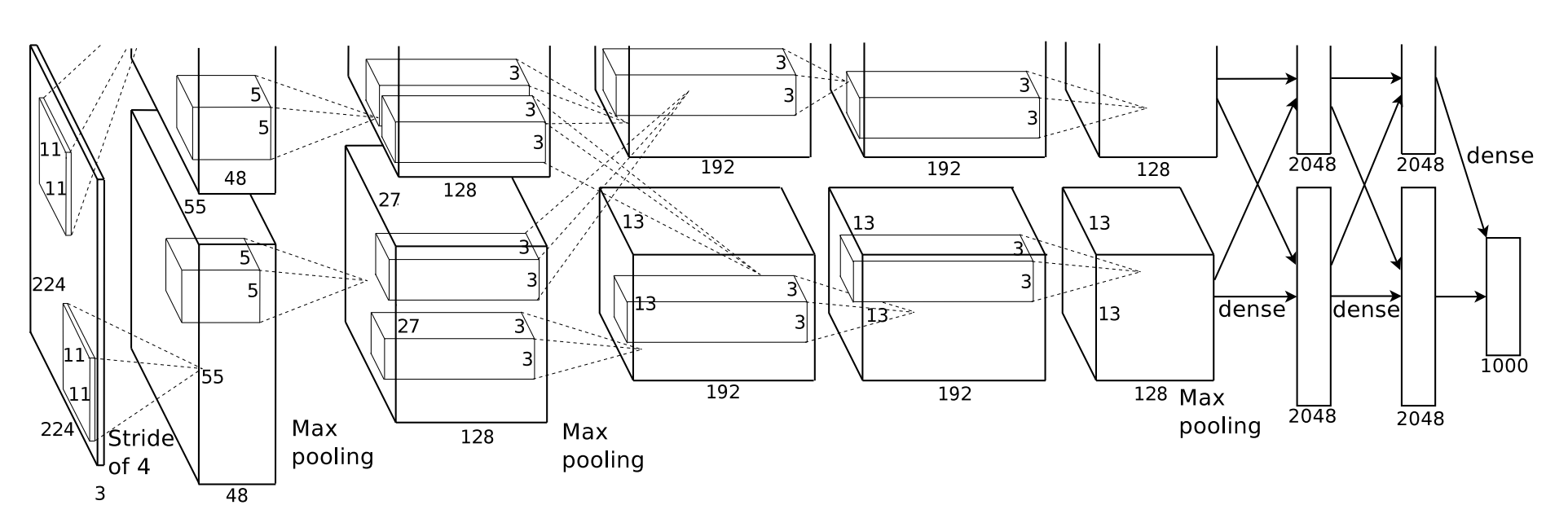}}
\vskip -0.1in
\caption{Neural network architecture of AlexNet~\cite{krizhevsky2012imagenet}, the winner of the ImageNet competition in 2012.}
\label{fig:alexnet}
\end{center}
\vskip -0.3in
\end{figure}

Furthermore, with all these advances, the impressive feats in deep learning involve sophisticated engineering efforts. For instance, the famous AlexNet~\cite{krizhevsky2012imagenet} (See Figure~\ref{fig:alexnet}), which put deep learning into the spotlight in the computer vision community after winning ImageNet in 2012, presented a carefully designed network architecture with a well-calibrated training procedure. Modern neural networks are often even more sophisticated, and require a pipeline that spans network architecture to delicate training schemes. Like many engineering projects, much labor and fine-tuning went into producing each result.

We believe that many of the limitations and side effects of deep learning stems from the fact that the current practice of deep learning is similar to the practice of engineering. The way we are building modern neural network systems is similar to the way we are building bridges and buildings, which are designs that are not adaptive. To quote Pickering, author of \textit{The Cybernetic Brain}~\cite{pickering2010cybernetic}: ``Most of the examples of engineering that come to mind are not adaptive. Bridges and buildings, lathes and power presses, cars, televisions, computers, are all designed to be indifferent to their environment, to withstand fluctuations, not to adapt to them. The best bridge is one that just stands there, whatever the weather.''

\begin{figure}[ht]
\vskip -0.0in
\begin{center}
\centerline{\includegraphics[width=0.383\columnwidth]{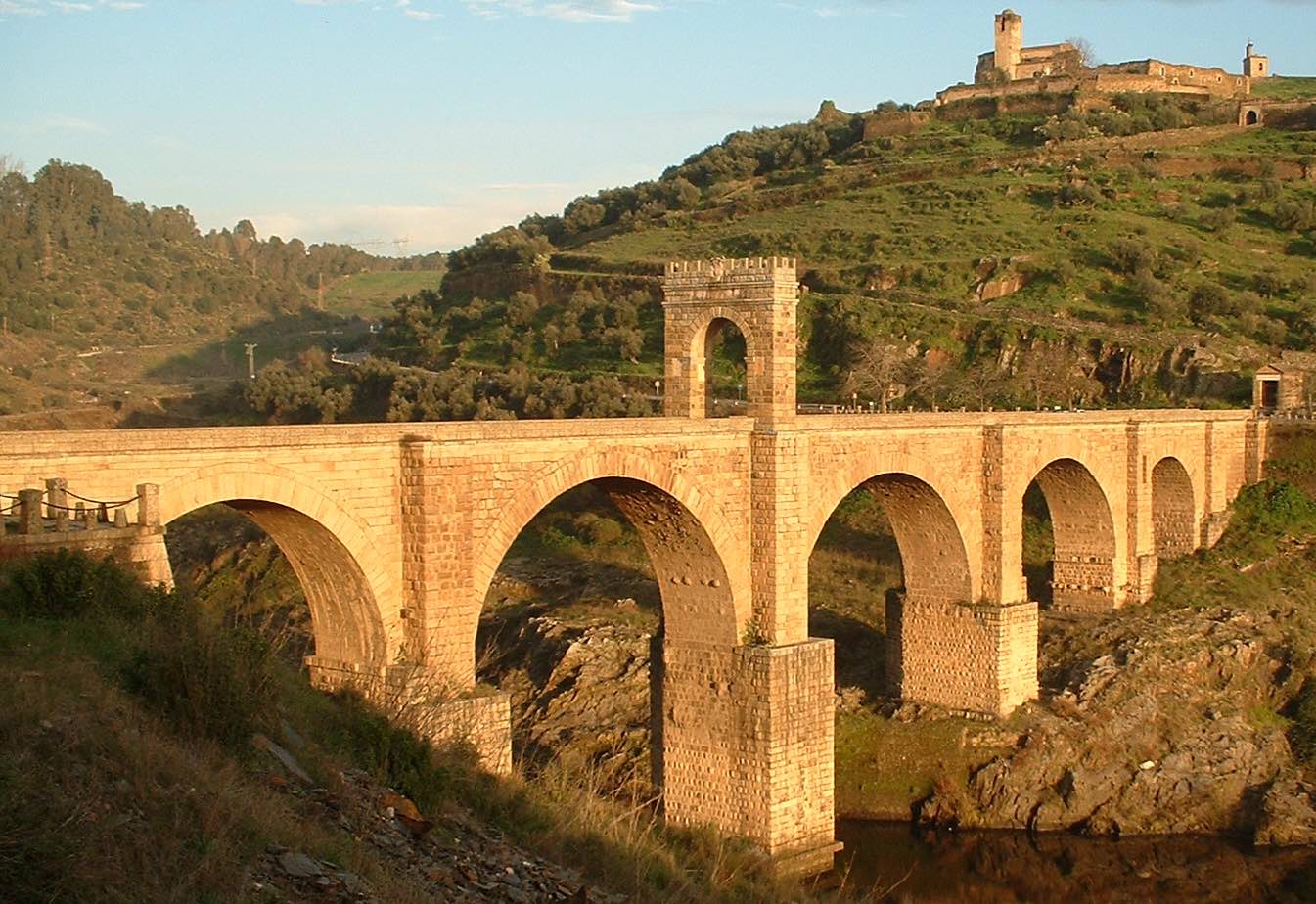}\includegraphics[width=0.327\columnwidth]{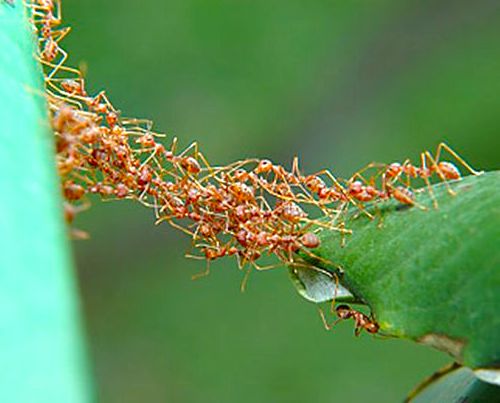}}
\vskip -0.00in
\caption{Left: Trajan's Bridge at Alcantara, built in 106 AD by Romans~\cite{alcantara2011}. Right: Army ants forming a bridge~\cite{janel2011}.}
\label{fig:intro_bridge}
\end{center}
\vskip -0.3in
\end{figure}

In natural systems, where collective intelligence plays a big role, we see \textit{adaptive} designs that emerge due to self-organization, and such designs are very sensitive and responsive to changes in the world around them. Natural systems adapt, and become part of their environment (See Figure~\ref{fig:intro_bridge} for an analogy).

As exemplified by the example of army ants collectively forming a bridge that adapts to its environment, collective behavior, commonly seen in nature, tends to produce systems that are adaptable, robust, and have less rigid assumptions about the environment configuration. Collective intelligence, as a field, studies the \textit{shared} intelligence that emerges from the interactions (such as collaboration, collective efforts, and competition) of many individuals. Within this field, ideas such as self-organization, emergent behavior, swarm optimization, and cellular automata were developed to model and explain complex systems. It is therefore natural to see these ideas incorporated into newer deep learning methods.

We do not believe that deep learning models have to be built in the same vein as bridges. As we will discuss later on, it didn't have to be this way. The reason why the deep learning field took this course could just be an accidental outcome in history. In fact, recently, several works have been addressing the limitations of deep learning by combining it with ideas from collective intelligence, from applying cellular automata to neural network-based image processing models~\cite{mordvintsev2020growing,randazzo2020self} to re-defining how problems in reinforcement learning can be approached using self-organizing agents~\cite{pathak2019learning,huang2020one,attentionneuron2021}. As we witness the continuous technological advances in parallel-computation hardware (which is naturally suited to simulate collective behavior, see Figure~\ref{fig:rudin2021learning} for an example), we can expect more works that incorporate collective intelligence into problems that have been traditionally approached with deep learning.

The goal of this review is to provide a high level survey of how ideas, tools, and insights central to the field of collective intelligence, most notably self-organization, emergence, and swarm models have impacted different areas of deep learning, ranging from image processing, reinforcement learning to meta-learning. We hope this review will provide some insights on future deep learning collective intelligence synergies, which we believe will lead to meaningful breakthroughs in both fields.

\section{Background: Collective Intelligence}

Collective intelligence (CI) is a term widely used in areas like sociology, business, communication and computer science. The definition of CI can be summarized as a form of distributed intelligence that is constantly enhancing and coordinating, with the goal of achieving better results than any individual of the group, through mutual recognition and enrichment of the individual~\cite{levy1997collective,leimeister2010collective}. The better results from CI are attributed to three factors: diversity, independence and decentralization~\cite{surowiecki2005wisdom,tapscott2008wikinomics}.


For our purposes, we view collective intelligence, as a field, to be the study of the \textit{group intelligence} that emerges from interactions (can be collaborative or competitive) between many individuals. This group intelligence is a product of \textit{emergence}, which occurs when the group is observed to have properties that the individuals that compose of the group do not have on their own, and \textit{emerge} only when the individuals of the group interact in a wider whole.

Examples of such systems are abounded in nature where complex global behaviors toward mutual goals emerge from simple local interactions/collaborations between individuals~\cite{deneubourg1989collective,toner2005hydrodynamics,sumpter2010collective,lajad2021young}.
In this review, we confine ourselves to be concerned with the \textit{simulation} of collective intelligence, rather than the analysis of CI observed in nature and society.
Decades of earlier work have also explored the simulation of collective behavior and to gather insights from such simulations.
\citet{mataric1993designing} investigated the use of physical mobile robots for studying social interactions leading to group behavior. They proposed a set of basic interactions (e.g., collision avoidance, following, flocking, etc) with the hope that these primitives would enable a group of autonomous agents to accomplish a common goal or to learn from each other. Inspired by group behaviors observed in real ant colonies, \citet{dorigo2000ant} posed stigmergy (a particular form of indirect communication used by social insects) as a distributed communication paradigm and showed how it inspired novel algorithms for solutions of distributed optimization and control problems. Moreover, \citet{schweitzer2003brownian} applied Brownian agent models in many different contexts. Combined with multi-agent systems and statistical approaches, the authors laid out a vision for a coherent framework for understanding complex systems.

While some of these earlier works led to the discovery of algorithms that are applicable to optimization problems (such as ant colony optimization for tackling the traveling salesman problem), many of these works aim to use these simulation models to \textit{understand} the emergent phenomenon of collective intelligence. This points to a fundamental difference between the goals of collective intelligence and artificial intelligence fields. In collective intelligence, the goal is to build models of complex systems that can help us explain and understand emergent phenomena, which may have applications to understand real systems in nature and society. Artificial intelligence (in particular, the field of machine learning), on the other hand, is concerned with optimization, classification, prediction, and \textit{solving} a problem.

The early works we mentioned did not fully leverage the modeling power of DL or the advancement of hardware development, but nonetheless are consistently demonstrating the incredible effects of CI. Namely, the systems are self-organizing, capable of optimization via swarm intelligence, present emergent behavior, etc. They suggest that concepts from CI are promising ideas that can be applied to DL to produce solutions that are robust, adaptable, and have less rigid assumptions about the environment configuration, which is the focus of this review.

\section{Historical Background: Cellular Neural Networks}

Ideas from complex systems such self-organization that were used to model and understand emergent and collective behavior have a long and interesting historical relationship with the development of artificial neural networks. While connectionism and artificial neural networks came about in the 1950s with the birth of artificial intelligence as a research field, our story begins in the 1970s, when a group of electrical engineers led by pioneer Leon Chua, started developing nonlinear circuits theory and applied it to computation. He is known for conceptualizing the Memristor in the 1970s (a device that has been implemented only recently), and devising the Chua circuit, one of the first circuits to exhibit chaotic behavior. In the 1980s, his group developed Cellular Neural Networks, which are computational systems that resemble cellular automata (CA), but use neural networks in place of the algorithmic cells typically seen in CA systems such as Conway's Game of Life~\cite{conway1970game} or elementary cellular automata rules~\cite{wolfram2002new}.

\begin{figure}[!htb]
\vskip -0.0in
\begin{center}
\centerline{\includegraphics[width=0.47\columnwidth]{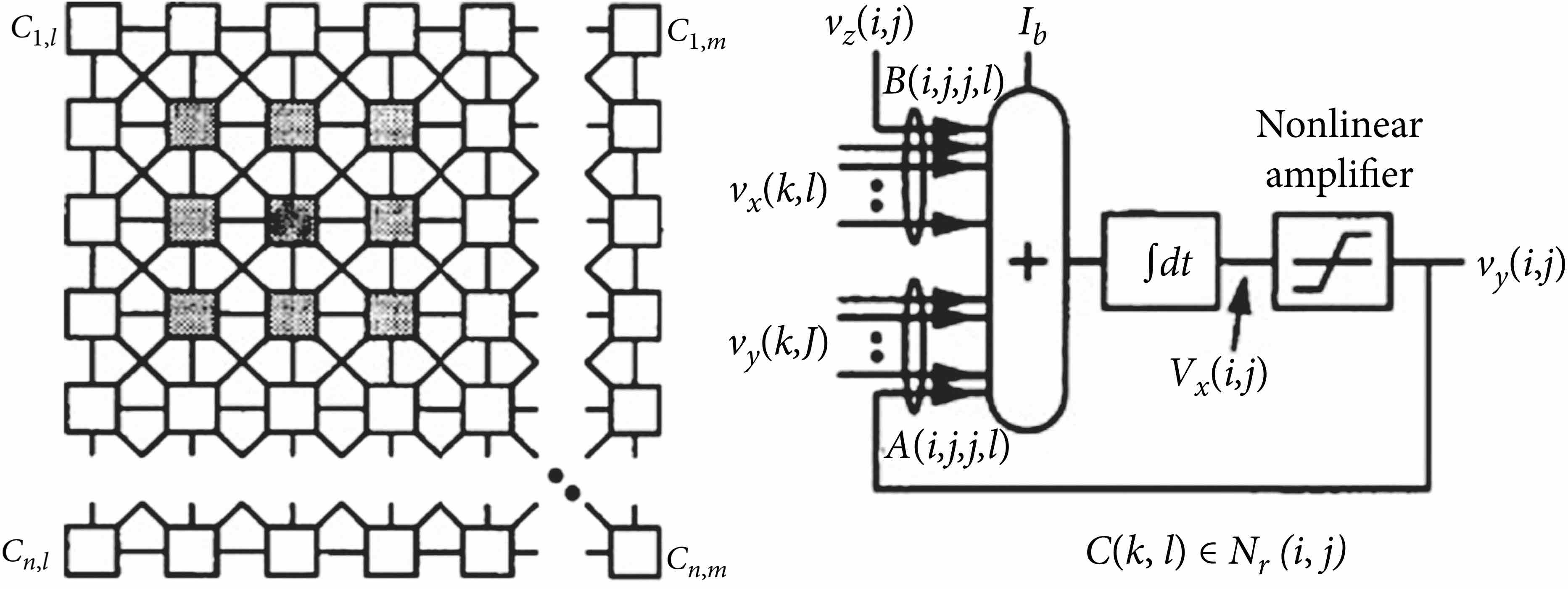}\includegraphics[width=0.53\columnwidth]{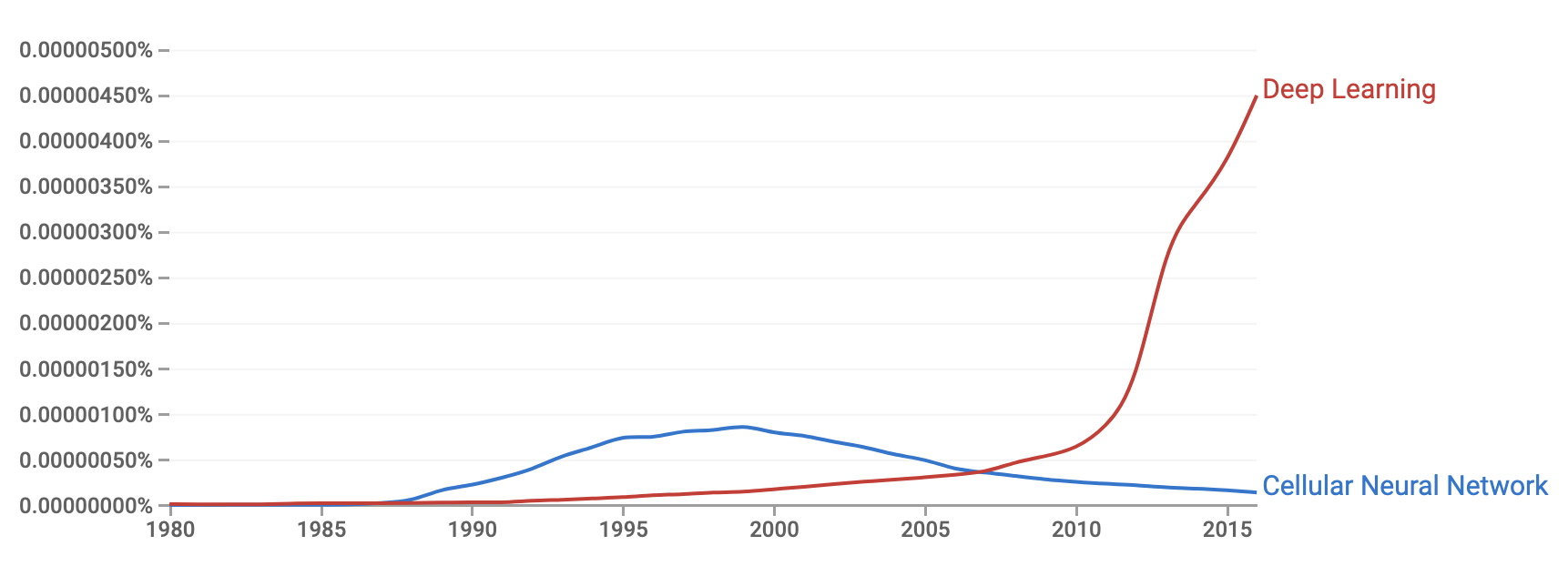}}
\vskip -0.00in
\caption{Left: Typical configuration of a 2D Cellular Neural Network~\cite{liu2020zagreb}. Right: \href{https://bit.ly/3sZlpyh}{Google trends} for the terms \textit{Deep Learning} and \textit{Cellular Neural Network} over time.}
\label{fig:cenn}
\end{center}
\vskip -0.2in
\end{figure}

Cellular Neural Networks (CeNNs)~\cite{chua1988cellulartheory,chua1988cellularapplication} are artificial neural networks where each \textit{neuron}, or \textit{cell}, can only interact with their immediate neighbors. In the most basic setting, the state of each cell is continuously updated using a nonlinear function of the states of its neighbors and itself. Unlike modern deep learning approaches which rely on digital, discrete-time computation, CeNNs are continuous-time systems that are usually implemented with non-linear \textit{analog} electronic components (See Figure~\ref{fig:cenn}, left), making them very fast. The dynamics of CeNNs rely on independent local processing of information and interaction between processing units, and like CAs, they also exhibit emergent behavior, and can be made to be Universal Turing Machines. However, they are vastly more general than discrete CAs and digital computers. Due to the continuous state space, CeNNs exhibit emergent behavior never seen before.~\cite{goras1995turing}

From the 1990s to mid 2000s, CeNNs became an entire subfield of AI research. Due to its powerful and efficient distributed computation, it found applications in image processing, texture analysis, and its inherent analog computation applied to solving PDEs and even modeling biological systems and organs.~\cite{chua2002cellular} There were thousands of peer-reviewed papers, textbooks, and an IEEE conference devoted to CeNNs, with many proposals to scale them up, stack them, combining them with digital circuits, and investigating different methods of training them (just like what we are currently seeing in deep learning). At least two hardware startups were formed to produce CeNN hardware and devices.

But in the latter half of the decade in the 2000s, they suddenly \textit{disappeared} from the scene! There is hardly any mention of Cellular Neural Networks in the AI community after 2006. And from the 2010s, GPUs took over as the predominant platform for neural network research, which led to the rebranding of artificial neural networks to \textit{deep learning}. See Figure~\ref{fig:cenn} (right) for a comparison of the trends over time.

No one can really pinpoint the exact reason for the demise of Cellular Neural Networks in AI research. Like the Memristor, perhaps CeNNs were ahead of its time. Or perhaps the eventual rise of consumer GPUs made it a compelling platform for deep learning. One can only imagine in a parallel universe where CeNN's analog computer chips had won the Hardware Lottery~\cite{hooker2020hardware}, the state of AI might be very different where the world and all of our devices are embedded with powerful distributed analog cellular automata.

However, one key difference between CeNNs and deep learning is accessibility, and in our opinion, this is the main reason it did not catch on. In the current deep learning paradigm, there is an entire ecosystem of tools designed to make it easy to train and deploy neural network models. It is also relatively straightforward to train the parameters of a neural network with deep learning frameworks by providing it with a dataset~\cite{chollet2015keras}, or a simulated task environment~\cite{hill2018stable}. Deep learning tools are designed to be used by anyone with a basic programming background. CeNNs, on the other hand, were designed for electrical engineers at a time when most EE students knew more about analog circuits than programming languages.

To illustrate this difficulty, ``\textit{training}'' a CeNN requires solving a system of at least nine ODEs to determine the coefficients that govern the analog circuits to define the behavior of the system! In practice, many practitioners needed to rely on a cookbook~\cite{chua2002cellular} of known solutions to problems and then manually adjust the solutions for new problems. Eventually, genetic algorithms (and early versions of backpropagation) have been proposed to train CeNNs~\cite{kozek1993genetic}, but they require simulation software to train and test the circuits, before deploying on an actual (and highly customized) CeNN hardware.

There are likely more lessons to be learned from Cellular Neural Networks. They were an immensely powerful hybrid of analog and digital computation that truly synthesized cellular automata with neural networks. Unfortunately, we probably only witnessed the very beginning of its full potential, before its demise. Ultimately, commodity GPUs and software tools that abstracted neural networks into simple Python code enabled deep learning to take over. Although CeNNs have faded away, concepts and ideas from complex systems, like CAs, self-organization and emergent behavior have not. Despite being limited to digital hardware, we are witnessing a resurgence of Collective Intelligence concepts in many areas of deep learning, from image generation, deep reinforcement learning, to collective and distributed learning algorithms. As we will see, these concepts are advancing the state of deep learning research by providing solutions to some limitations and restrictions of traditional artificial neural networks. 


\begin{figure}[!htb]
\vskip -0.0in
\begin{center}
\centerline{\includegraphics[width=0.6\columnwidth]{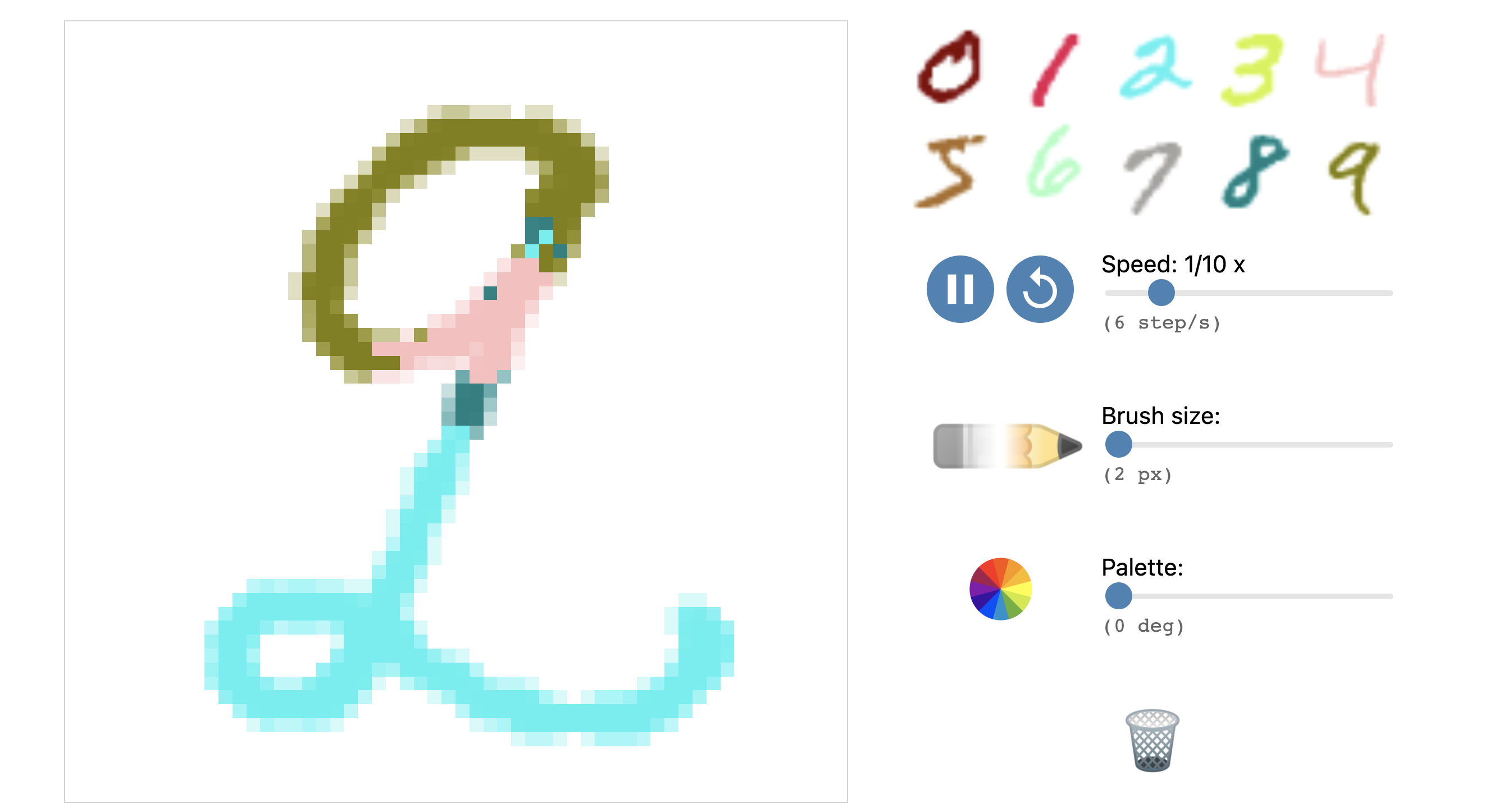}}
\vskip -0.00in
\caption{A Neural Cellular Automata trained to recognize MNIST digits created by Randazzo et al.~\cite{randazzo2020self} is also available as an interactive web demo. Each cell is only allowed to see the contents of a single pixel and communicate with its neighbors. Over time, a consensus will be formed as to which digit is the most likely pixel, but interestingly, disagreements may result depending on the location of the pixel where the prediction is made.}
\label{fig:randazzo2020self}
\end{center}
\vskip -0.55in
\end{figure}

\section{Collective Intelligence for Deep Learning}

Collective intelligence naturally arises from the interaction of multiple individuals in a network, and it is no surprise to also see self-organizing behaviors naturally emerge from artificial neural networks. This is especially true when we employ repeated computation of identical modules with identical weight parameters across the network. For example, Gilpin~\cite{gilpin2019cellular} observed the close connection between cellular automata and convolutional neural networks (CNNs), a type of neural network often used in image processing that applies the same weights (or \textit{filters}) to all of its inputs. In fact, they show that any CA can be represented with a certain kind of CNN, and with an elegant demonstration of Conway's Game of Life~\cite{conway1970game} in a CNN, illustrating that in certain settings, CNNs can exhibit interesting self-organizing behaviors. Recently, several works such as Mordintsev et al.~\cite{mordvintsev2020growing} that we will discuss later have exploited the self-organizing properties of CNN, and have developed neural network-based cellular automata for applications such as image regeneration.

Other types of neural network architectures, such as Graph Neural Networks~\cite{wu2020comprehensive,sanchez2021gentle,daigavane2021understanding} explicitly target self-organizing as a central feature, modeling the behavior of each node of a graph as identical neural network modules that pass messages to their neighbors defined by the edges of a graph. GNNs have been traditionally used to analyze graph domains such as social networks and molecular structures. Recent work~\cite{grattarola2021learning} has also demonstrated the ability of GNNs to learn rules for established CA systems such as Voronoi diagrams, or the flocking behavior of swarms~\cite{schoenholz2020jax}. As we will discuss later, the self-organizing properties of GNNs have recently been applied to the deep reinforcement learning domain, creating agents with far superior generalization capabilities.

We have identified four areas of deep learning that have started to incorporate ideas related to collective intelligence: (1) Image Processing, (2) Deep Reinforcement Learning, (3) Multi-agent Learning, and (4) Meta-Learning. We will discuss each area in detail and provide examples in this section.

\subsection{Image Processing}

Implicit relationships and recurring patterns in nature (such as texture and scenery) can benefit from employing approaches from cellular automata in learning alternative representations of natural images.
Like CeNNs, the \textit{Neural Cellular Automata} (\textit{neural CA}) model proposed by Mordvintsev et al.~\cite{mordvintsev2020growing} treated each individual pixel of an image as a single neural network cell. The networks are trained to predict its color based on the states of its immediate neighbors, thereby developing a model of morphogenesis for image generation. They demonstrated that it was possible to train neural networks to reconstruct entire images this way, even when each cell lacks information about its location and rely only on local information from its neighbors. This approach enabled the generation algorithm to be resistant to noise, and moreover, allowed images to \textit{regenerate} when damaged. An extension of neural CA~\cite{randazzo2020self} enabled individual cell to perform image classification tasks, such as handwritten digit classification (MNIST) by only examining the contents of a single pixel, and passing a message on to the cell's immediate neighbors (See Figure~\ref{fig:randazzo2020self}). Over time, a consensus will be formed as to which digit is the most likely pixel, but interestingly, disagreements may result depending on the location of the pixel, especially if the image is intentionally drawn to represent different digits.

\begin{figure}[ht]
\vskip -0.0in
\begin{center}
\centerline{\includegraphics[width=0.33\columnwidth]{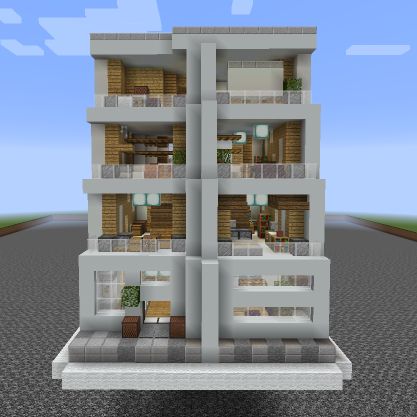}\includegraphics[width=0.33\columnwidth]{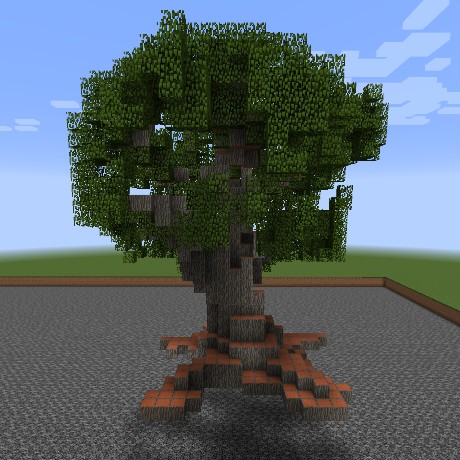}\includegraphics[width=0.33\columnwidth]{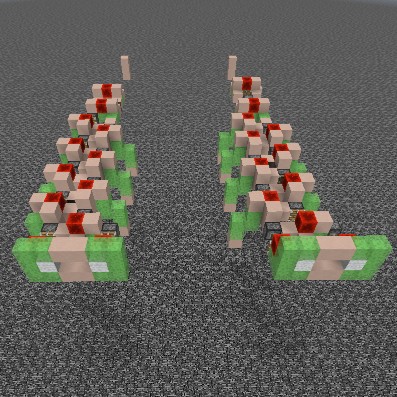}}
\vskip -0.00in
\caption{Neural CAs have also been applied to the regeneration of Minecraft entities. Sudhakaran et al.~\cite{sudhakaran2021growing}'s formulation enabled the regeneration of not only Minecraft buildings, trees, but also simple functional machines in the game such as worm-like creatures that can even regenerate into two distinct creatures when cut in half.}
\label{fig:sudhakaran2021growing}
\end{center}
\vskip -0.3in
\end{figure}

The regeneration with neural CA has been explored beyond 2D images. In a later work, Zhang et al.~\cite{zhang2021learning} employed a similar approach to 3D voxel generation. This is particularly useful for high-resolution 3D scanning, where 3D shape data is often described with sparse and incomplete points. Using generative cellular automata they can recover full 3D shapes from only a partial set of points. This approach is also applicable outside of pure generative domains, and can also be applied to the construction of artificial agents in active environments such as Minecraft. Sudhakaran et al.~\cite{sudhakaran2021growing} trained neural CAs to grow complex entities from Minecraft such as castles, apartment blocks, and trees, some of which are composed of thousands of blocks. Aside from regeneration, their system is able to regrow parts of simple functional machines (such as a virtual creature in the game), and they demonstrate a morphogenetic creature grow into two distinct creatures when cut in half in the virtual world (See Figure~\ref{fig:sudhakaran2021growing}).

Cellular Automata is also naturally applicable to provide visual interpretation for images. Qin at al.~\cite{qin2018hierarchical} examined the use of a Hierarchical CA model for visual saliency, to identify items in an image that stand out. By getting a CA to operate on visual features extracted from a deep neural network, they were able to iteratively construct multi-scale saliency maps of the image, with the final image being close to the target items. Sandler et al.~\cite{sandler2020image} later investigated the use of CA for the task of image segmentation, an area where deep learning enjoys tremendous success. They demonstrated the viability of performing complex segmentation tasks using CAs with relatively simple rules (with as little as 10K neural network parameters), with the advantage of the approach being able to scale up to incredibly large image sizes, a challenge for traditional deep learning models with \textit{millions} or even \textit{billions} of model parameters, which are bounded by GPU memory.

\subsection{Deep Reinforcement Learning}

The rise in deep learning gave birth to the use of deep neural networks for reinforcement learning, or \textit{deep reinforcement learning} (Deep RL), equipping reinforcement learning agents with modern neural networks architectures that can address more complex problems, such as high dimensional continuous control or vision-based tasks from pixel observations. While Deep RL shares successful characteristics with deep learning, in that employing sufficient computation resources will generally lead to the solution of a target \textit{training} task to be found. But like deep learning, Deep RL has its share of limitations. Agents trained to perform a particular task often fail when the task is slightly altered. Furthermore, neural network solutions generally only work for a specific morphology with well-defined input and output mappings. For instance, a locomotion policy trained for a 4-legged ant might not work for a 6-legged one, and a controller that expects to receive 10 inputs won't work if you give it 5, or 20 inputs.

\begin{figure}[ht]
\vskip -0.15in
\begin{center}
\centerline{\includegraphics[width=0.4\columnwidth]{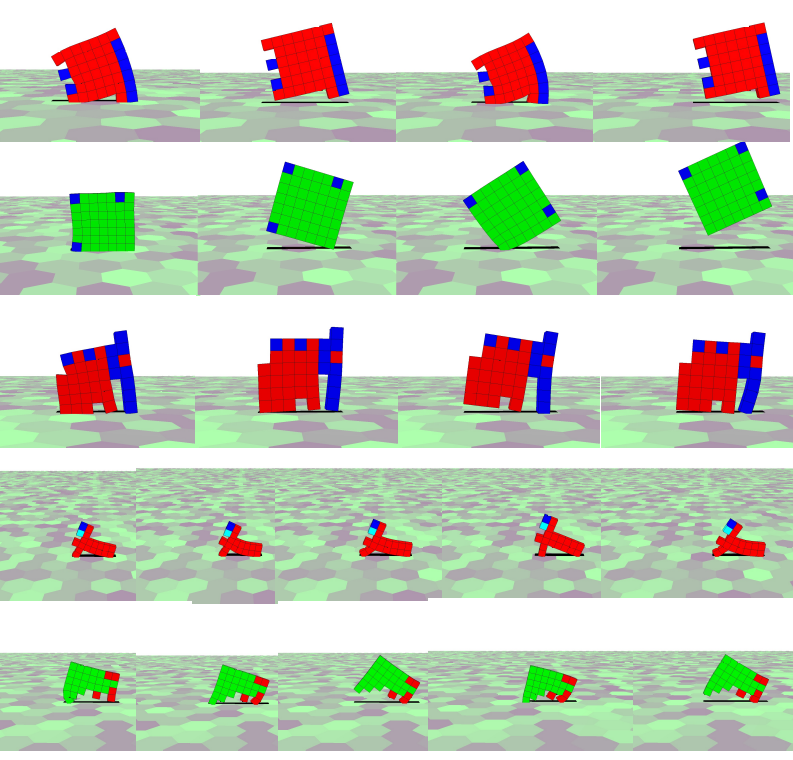}\includegraphics[width=0.4\columnwidth]{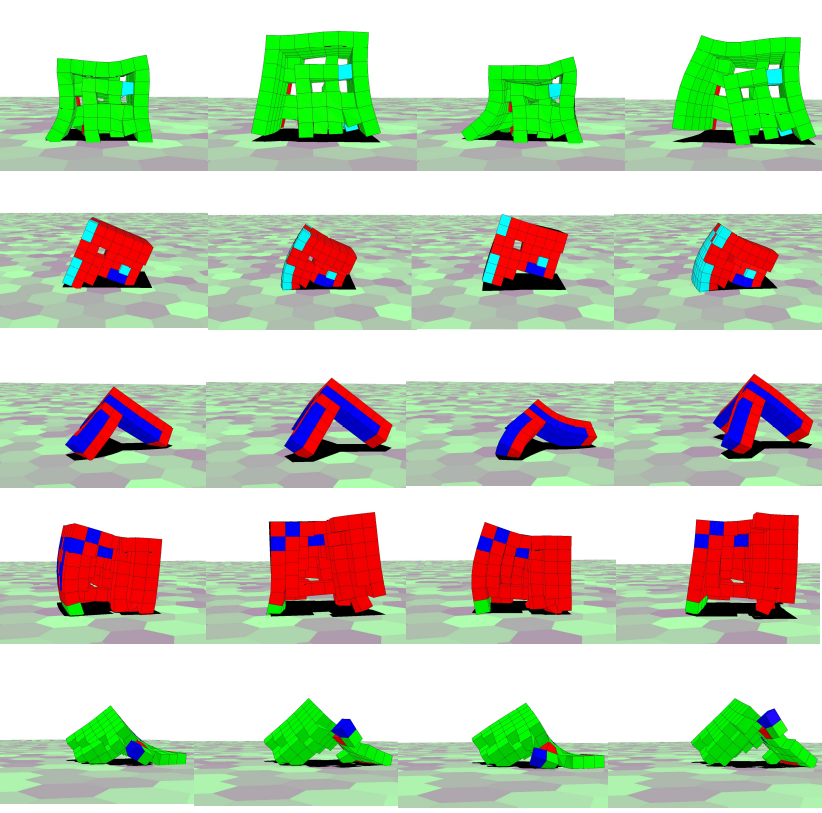}}
\vskip -0.00in
\caption{Examples of soft-bodied robot simulation in 2D and 3D. Each cell represents an individual neural network with local sensory functions that produce local actions, including communicating to neighboring cells. Training these systems to perform various locomotion tasks involve not only training the neural networks, but also the design and placement of the soft cells that form the agent's morphology. Figure from Horibe et al.~\cite{horibe2021regenerating}}
\label{fig:horibe2021regenerating}
\end{center}
\vskip -0.3in
\end{figure}

The evolutionary computation community started approaching some of these challenges earlier on, by incorporating modularity~\cite{schilling2000toward,schilling2001use} in the evolutionary process that govern the design of artificial agents. Having agents that are composed of identical but independent modules foster self-organization via local interactions between the modules, enabling systems that are robust to changes in the agent's morphology, an essential requirement in evolutionary systems. These ideas have been presented in the literature of work on soft-bodied robotics~\cite{cheney2014unshackling}, where robots consist of a grid of voxel cells--each controlled by an independent neural network with local sensory function that can produce a localized action. Through message passing, the group of cells that make up the robot are able to self-organize and perform a range of locomotion tasks (See Figure~\ref{fig:horibe2021regenerating}). Later work~\cite{joachimczak2016artificial} even proposes incorporating metamorphosis in the evolution of the placement of the cells to produce configurations robust to a range of environments.

\begin{figure}[ht]
\vskip -0.0in
\begin{center}
\centerline{\includegraphics[width=0.8\columnwidth]{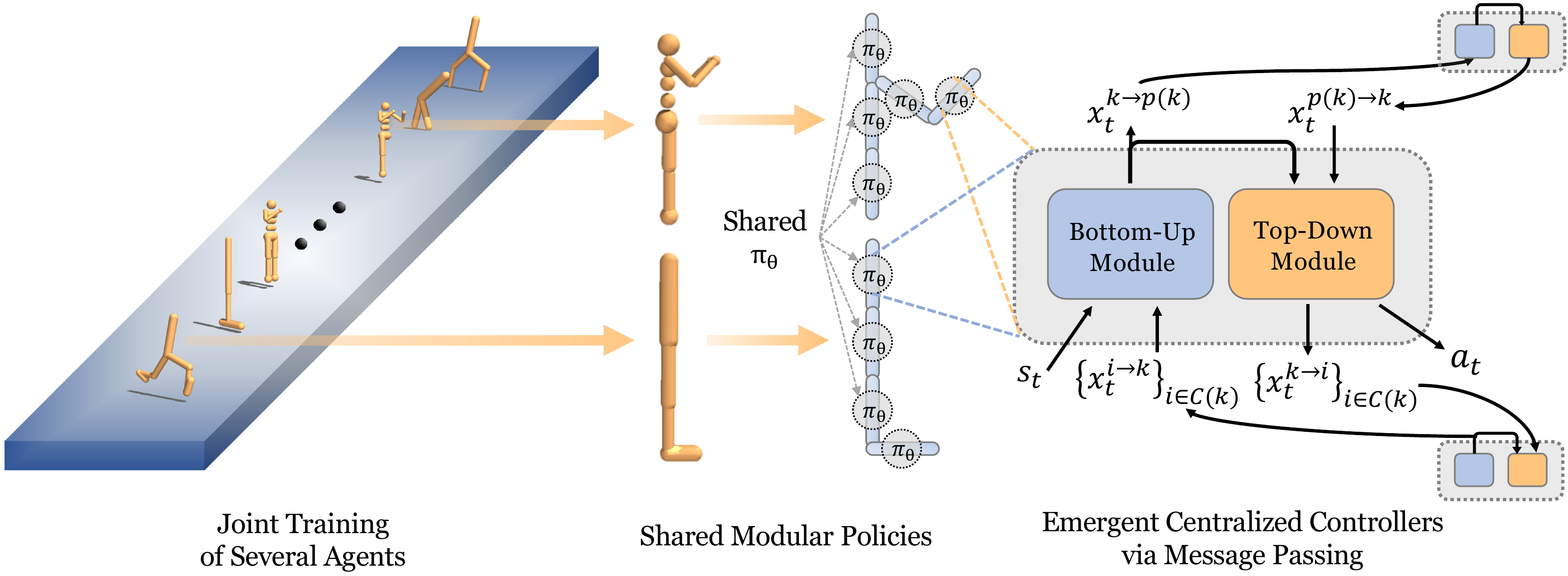}}
\vskip -0.00in
\caption{Traditional RL methods train a specific policy for a particular robot with a fixed morphology. But recent work, like the one shown here by Huang et al.~\cite{huang2020one} attempts to train a \textit{single} modular neural network responsible for controlling a single part of a robot. The global policy of each robot is thus the result of coordination of these identical modular neural networks. They show that such a system can generalize across a variety of different skeletal structures, from hoppers to quadrupeds, and even to some unseen morphologies.}
\label{fig:huang2020one}
\end{center}
\vskip -0.35in
\end{figure}

Recently, soft-bodied robots have even been combined with the neural CA approach discussed earlier to enable these robots to regenerate themselves.~\cite{horibe2021regenerating} To bridge the gap between policy optimization (where the goal is to find the best parameters of the policy neural network) usually done in the Deep RL community and the type of morphology-policy co-evolution (where \textit{both} the morphology and the policy neural network is optimized together) work done in the soft-bodied literature, Bhatia et al.~\cite{bhatia2021evolution} has recently developed an OpenAI Gym-like~\cite{brockman2016openai} environment called \textit{Evolution Gym}, a benchmark for developing and comparing algorithms for co-optimizing design and control, which provided an efficient soft-bodied robot simulator written in C++ with a Python interface.

Modular, decentralized self-organizing controllers have also started to be explored in the Deep RL community. Wang et al.~\cite{wang2018nervenet} and Huang et al.~\cite{huang2020one} explored the use of modular neural networks to control each individual actuator of a simulated robot for continuous control. They expressed a global locomotion policy as a collection of modular neural networks (in the case of Huang et al.~\cite{huang2020one}, identical networks) that correspond to each of the agent's actuators, and trained the system using RL. Like soft-bodied robots, every module is only responsible for controlling its corresponding actuator and receives information from only its local sensors (See Figure~\ref{fig:huang2020one}). Messages are passed between neighboring modules, propagating information between distant modules. They show that a single modular policy can generate locomotion behaviors for several distinct robot morphologies, and show that the policies generalize to variations of the morphologies not seen during training, such as creatures with extra legs. As in the case of soft-bodied robots, these results also demonstrate the emergence of centralized coordination via message passing between decentralized modules that are collectively optimizing for a shared reward.

\begin{figure}[ht]
\vskip -0.0in
\begin{center}
\centerline{\includegraphics[width=0.8\columnwidth]{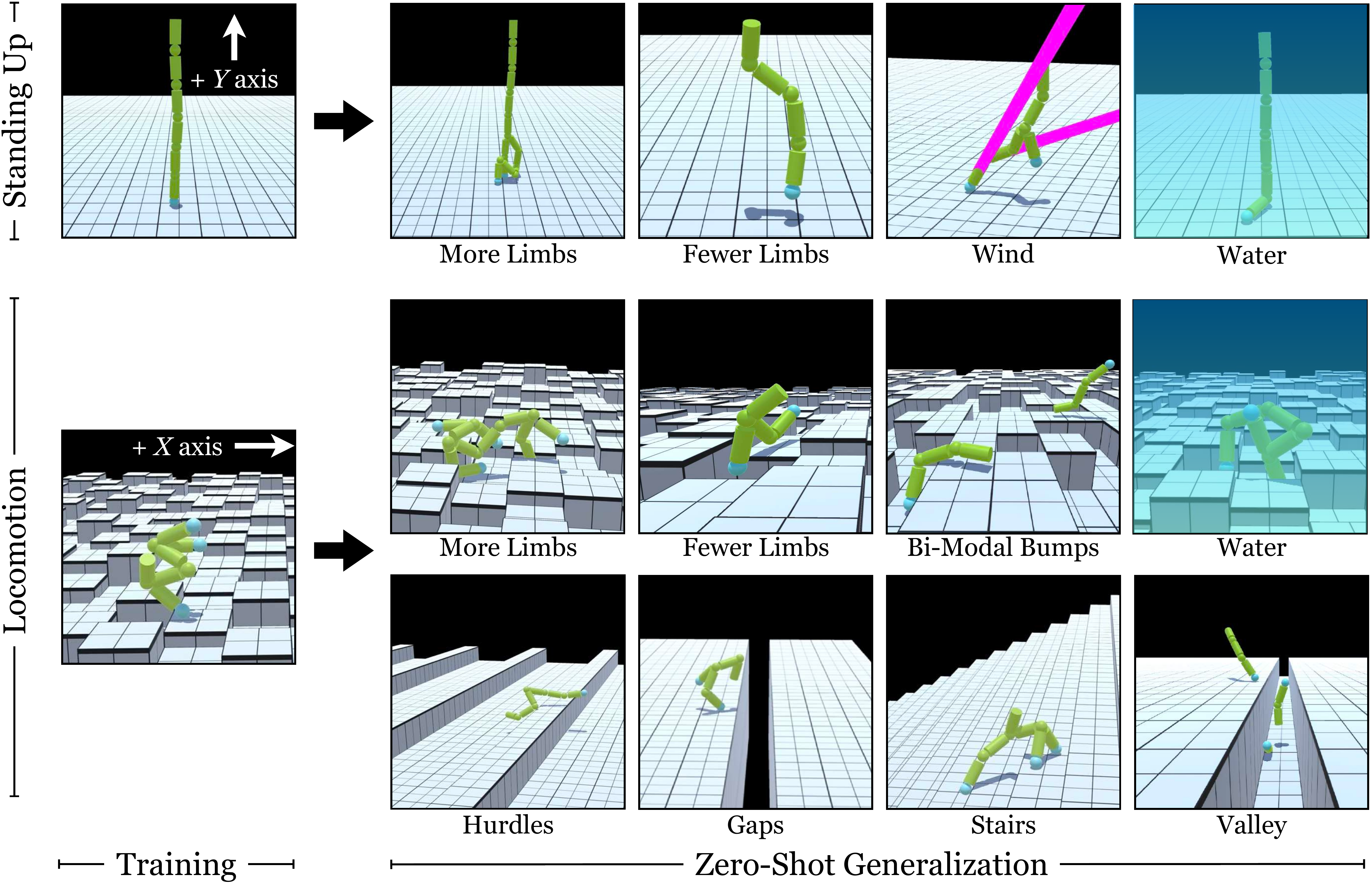}}
\vskip -0.00in
\caption{Self-organization also enables systems in RL environments to self-configure its own design for a given task. Pathak et al.~\cite{pathak2019learning} explored such dynamic and modular agents and showed that they can generalize to not only unseen environments, but also to unseen morphologies composed of additional modules.}
\label{fig:pathak2019learning}
\end{center}
\vskip -0.35in
\end{figure}

The aforementioned work hints at the power of embodied cognition, which emphasizes the role of the agent's body in generating behavior. Although the focus of much of the work in Deep RL is in learning neural network policies for an agent with a fixed design (e.g., a bipedal robot, humanoid, or robot arm), embodied intelligence is an area that is gathering interest in the sub-field~\cite{ha2018designrl,pathak2019learning}. Inspired by previous work on self-configuring modular robots~\cite{stoy2010self,rubenstein2014programmable,hamann2018swarm}, Pathak et al.~\cite{pathak2019learning} investigates a collection of primitive agents that learn to self-assemble into a complex body while also learning a local policy to control the body without an explicit centralized control unit. Each primitive agent (which consists of a limb and a motor) can link up with nearby agents, allowing for complex morphologies to emerge. Their results show that these dynamic and modular agents are robust to changes in conditions and the policies can generalize to not only unseen environments, but also to unseen morphologies consisting of a greater number of modules. We note that these ideas can be used to allow general DL systems (not confined to RL) to have more flexible architectures that can even \textit{learn} machine learning algorithms, and we will discuss this later on in the Meta-Learning section.

\begin{figure}[ht]
\vskip -0.0in
\begin{center}
\centerline{\includegraphics[width=0.53\columnwidth]{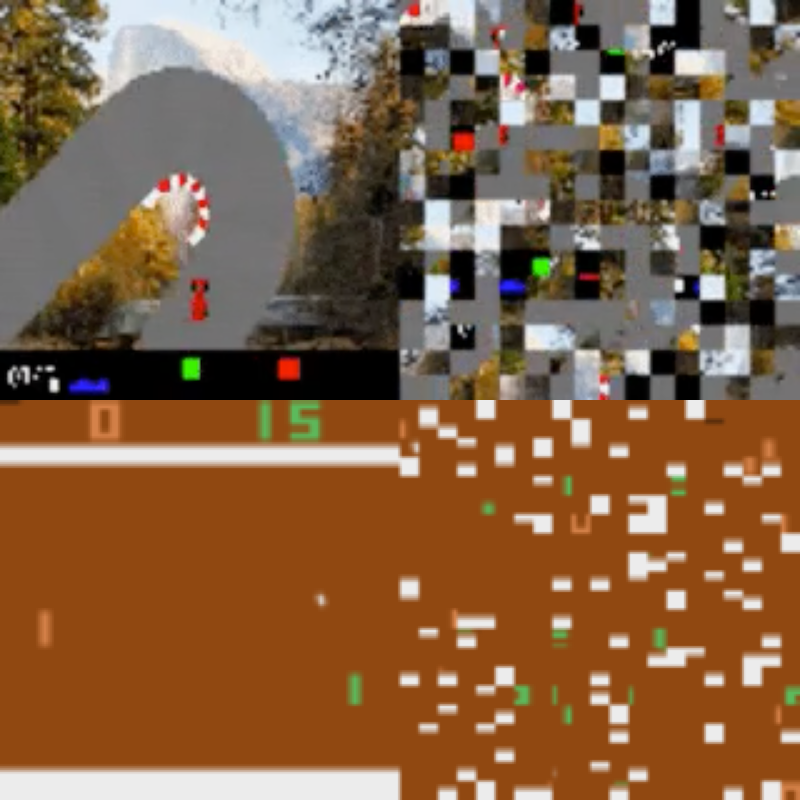}\includegraphics[width=0.47\columnwidth]{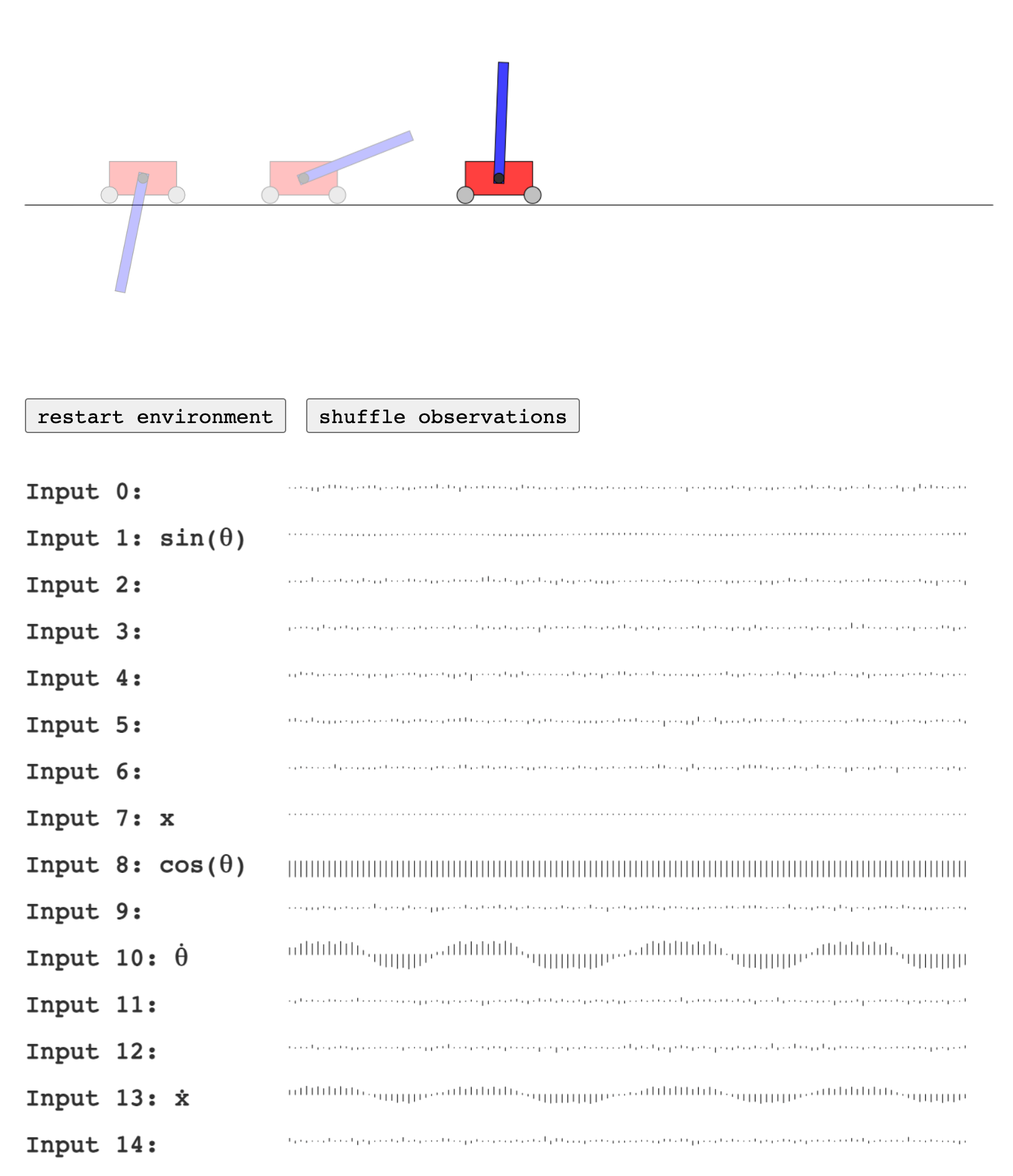}}
\vskip -0.00in
\caption{Using the properties of self-organization and attention, Tang and Ha~\cite{attentionneuron2021} investigated RL agents that treat their observations as an arbitrarily ordered, variable-length list of sensory inputs. They partition the input in visual tasks such as CarRacing and Atari Pong~\cite{brockman2016openai,attentionagent2020} into a 2D grid of small patches, and shuffled their ordering (Left). They also added many additional redundant noisy input channels in continuous control tasks~\cite{learningtopredict2019} in a shuffled order (Right), where the agent has to learn to identify which inputs are useful. Each sensory neuron in the system receives a stream of a particular input, and through coordination, must complete the task at hand.}
\label{fig:attentionneuron2021}
\end{center}
\vskip -0.4in
\end{figure}

Aside from adapting to changing morphologies and environments, self-organizing systems can also adapt to changes in their sensory inputs. Sensory substitution refers to the brain’s ability to use one sensory modality (e.g., touch) to supply environmental information normally gathered by another sense (e.g., vision). However, most neural networks are not able to adapt to sensory substitutions. For instance, most RL agents require their inputs to be in an exact, pre-specified rigid format, otherwise they will fail. In a recent work, Tang and Ha~\cite{attentionneuron2021} explored permutation invariant neural network agents that require each of their sensory neurons (receptors that receive sensory inputs from the environment) to deduce the meaning and context of its input signal, rather than explicitly assume a fixed meaning. They demonstrate that these sensory networks can be trained to integrate information received locally, and through communication between them using an attention mechanism, can collectively produce a globally coherent policy. Moreover, the system can still perform its task even if the ordering of its sensory inputs (represented as real-valued numbers) is randomly permuted several times during an episode. Their experiments show that such agents are robust to observations that contain many additional redundant or noisy information, or observations that are corrupt and incomplete.

\subsection{Multi-agent Learning}

Collective intelligence can be viewed at several different scales.
The brain can be viewed as a network of individual neurons functioning collectively. Each organ can be viewed as a collection of cells performing a collective function. Individual animals can be viewed as a collection of organs working together. As we zoom out further, we can also look at human intelligence beyond biology and see human civilization as a collective intelligence solving (and producing) problems that are beyond the capabilities of a single person. As such, while in the previous section, we discussed several works that leverage the power of collective intelligence to essentially decompose a single RL agent into a collection of smaller RL agents working together towards a collective goal, resembling a model of collective intelligence at the biological level, we can also view multi-agent problems as a model of collective intelligence at the societal level.

A major focus of the collective intelligence field is to study the group intelligence and behaviors emerged from a \textit{large} collection of individuals, whether in humans~cite{tapscott2008wikinomics}, animals~\cite{sumpter2010collective} insects~\cite{dorigo2000ant,seeley2010honeybee}, or artificial swarm robots~\cite{hamann2018swarm,rubenstein2014programmable}. This focus has clearly been missing in the Deep RL field. While \textit{multi-agent reinforcement learning} (MARL) is a well-established \text{branch} of Deep RL, most learning algorithms and environments proposed have targeted a relatively small number of agents~\cite{foerster2016learning,oroojlooyjadid2019review}, and thus not sufficient to study the emergent properties from large populations. In the most common MARL environments~\cite{resnick2018pommerman,baker2019emergent,jaderberg2019human,terry2020pettingzoo}, ``multi-agent'' simply means 2 or 4 agent trained to perform a task by means of \textit{self-play}~~\cite{bansal2017emergent,liu2019emergent,slimevolleygym2020}. Collective intelligence observed in nature or in society, however, relies on a much larger number of individuals than typically studied in MARL, involving population sizes from thousands to million. In this section, we will discuss recent works from the MARL sub-field of Deep RL that had been inspired by collective intelligence (as their authors have even noted in their publications). Unlike most MARL works, these work started to employ large population of agents (each enabled by a neural network), from thousands to millions, in order to truly study their emergent properties at the macro level (1000+ agents), rather than at the micro-level (2-4 agents).

\begin{figure}[ht]
\vskip -0.0in
\begin{center}
\centerline{\includegraphics[width=0.65\columnwidth]{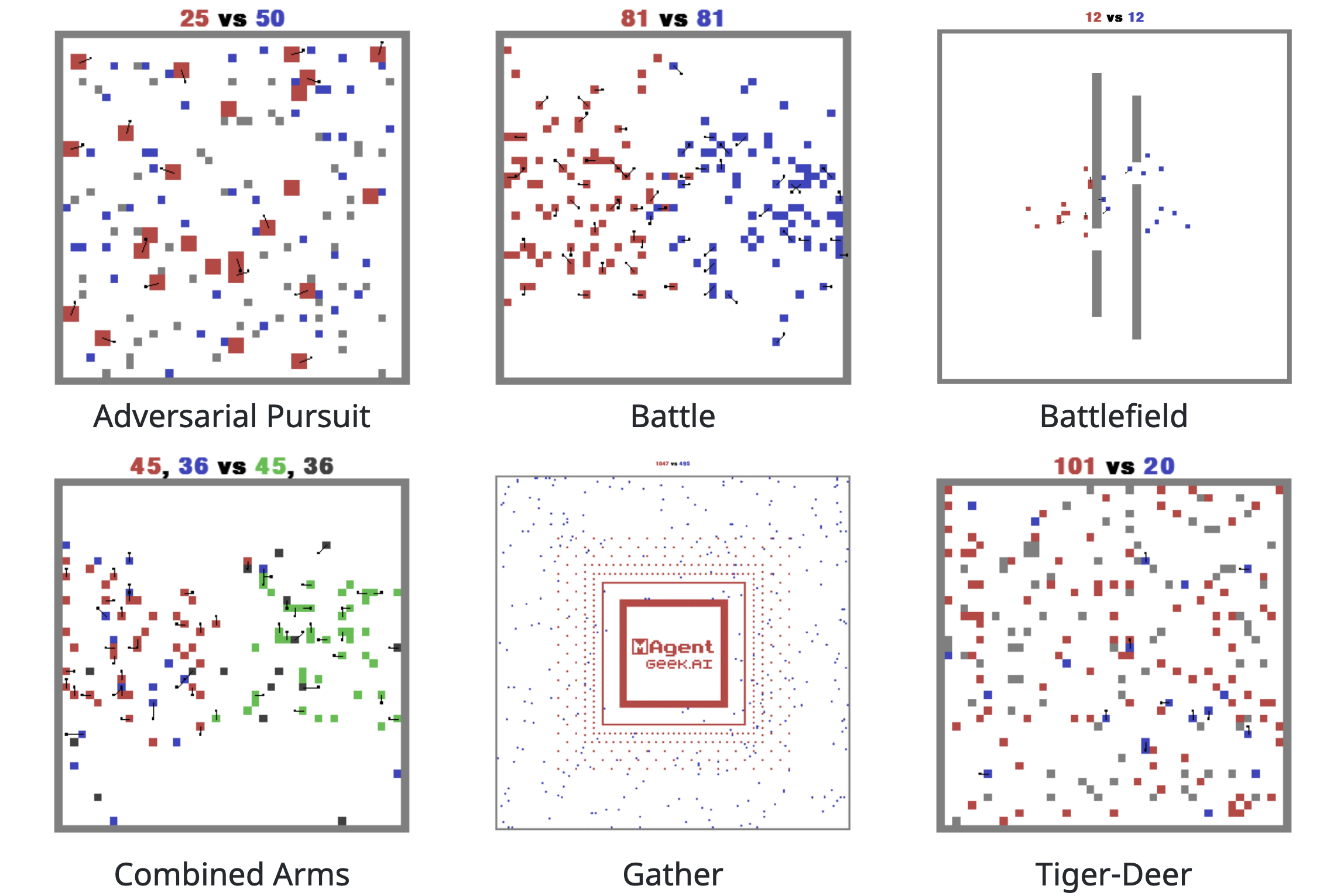}}
\vskip -0.00in
\caption{MAgent~\cite{zheng2018magent} is a set of environments where large numbers of pixel agents in a gridworld interact in battles or other competitive scenarios. Unlike most platforms that focus on RL research with a single agent or only few agents, their aim is to support RL research that scales up to millions of agents. The environments in this platform are now maintained as part of the PettingZoo~\cite{terry2020pettingzoo} open-source library for multi-agent RL research.}
\label{fig:zheng2018magent}
\end{center}
\vskip -0.4in
\end{figure}

Recent advances in Deep RL have demonstrated the capabilities of simulating thousands of agents in complex 3D simulation environments using only a single GPU~\cite{heiden2021neuralsim,rudin2021learning}. A key challenge is in approaching the problem of multi-agent learning at a much larger scale, leveraging such advances in parallel computing hardware and distributed computation, with the goal of training millions of agents. In this section, we will example recent attempts at training a massive number of agents that interact in a collective setting.

Rather than focusing on realistic physics or environment realism, Zheng. et al.~\cite{zheng2018magent} developed a platform called \textit{MAgent}, a simple grid-world environment that can allows millions of neural network agents. Their focus is on scalability, and they demonstrate that MAgent can host up to a million agents on a single GPU (in 2017). Their platform supports interactions among the population of agents, and facilitates not only the study of learning algorithms for policy optimization, but more critically, enables the study of social phenomena emerging from the millions of agents in an AI society, including the emergence of languages and societal hierarchy structures that may have emerged. Environments can be built using scripting, and they have provided examples such as predator-prey simulations, battlefields, adversarial pursuit, supporting different species of distinct agents that may exhibit different behaviors.

MAgent inspired many recent applications, including multi-agent driving~\cite{peng2021learning}, which looks at emergent behavior of entire populations of driving agents to optimize the driving policies which not only affect a single car, but aim to improve the safety of the population as a whole. These directions are good examples that demonstrate the difference between problems framed for deep learning (finding a driving policy for a single car) versus problems in collective intelligence (finding a driving policy for the entire population).

\begin{figure}[ht]
\vskip -0.0in
\begin{center}
\centerline{\includegraphics[width=0.65\columnwidth]{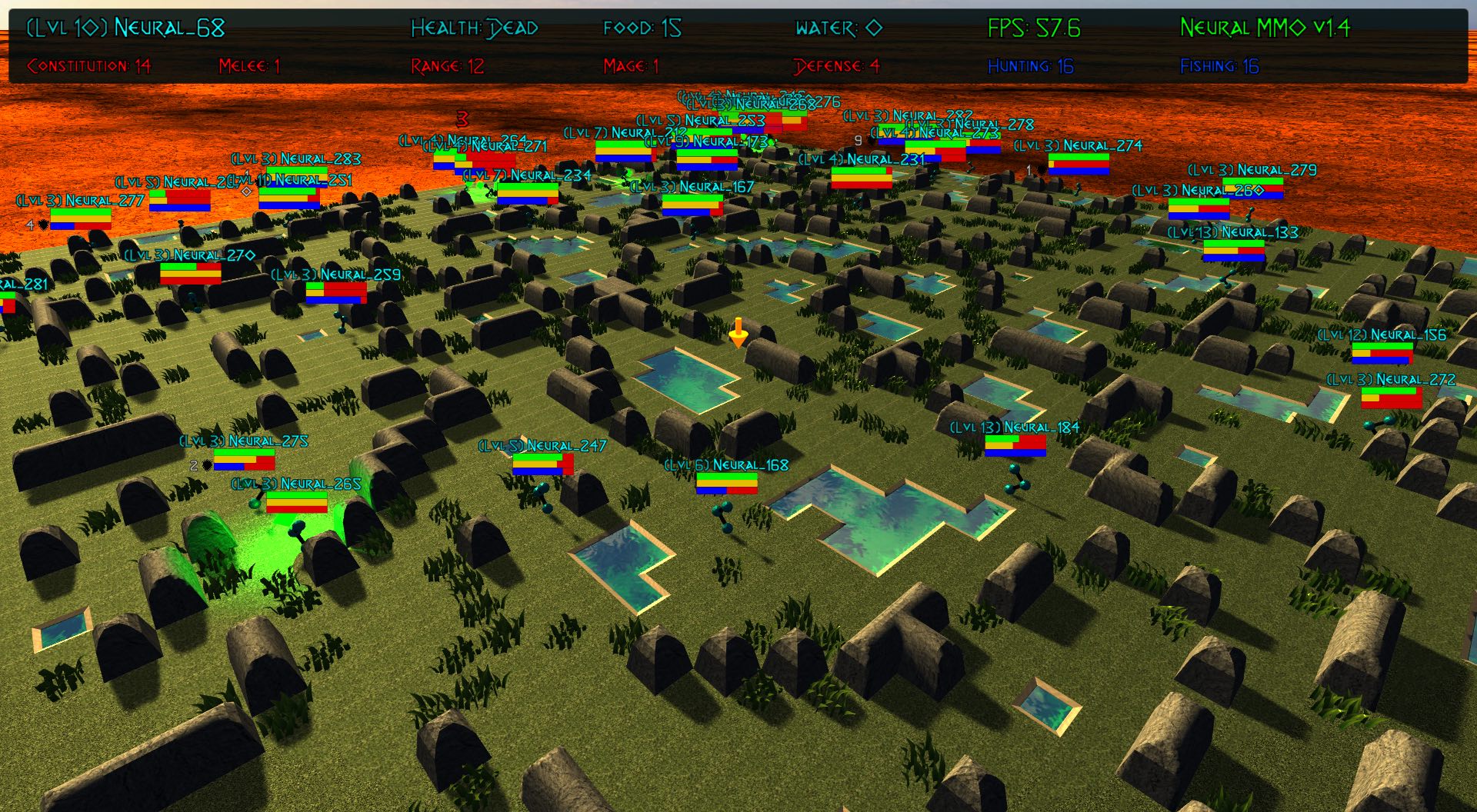}}
\vskip -0.00in
\caption{Neural MMO~\cite{suarez2021neural} is a platform that simulates populations of agents in procedurally generated virtual worlds to support multi-agent research while keeping its requirements computationally accessible. Users select from a set of provided game systems to create environments for their specific research problems–-with support for up to a thousand agents and one square kilometer maps over several thousand time steps. The project is under active development, with extensive documentation and tools that provide logging, and visualization tools for researchers. As of writing, this platform is to be demoed at the NeurIPS 2021 conference.}
\label{suarez2021neural}
\end{center}
\vskip -0.4in
\end{figure}

Inspired by the game genre of MMORPGs (Massively Multiplayer Online Role-Playing Games, aka MMOs), Neural MMO~\cite{suarez2021neural} is an AI research environment that supports a large number of artificial agents that have to compete for finite resources in order to survive. As such, their environment enables large-scale simulation of multi-agent interactions that requires agents to learn combat and navigation policies alongside other agents in a large population all attempting to do the same. Unlike most MARL environments, each agent is allowed to have their own distinct set of neural network weights, which has been a technical challenge in terms of memory consumption. Preliminary experimental results in early versions of the platform~\cite{suarez2019neural} demonstrated agents with distinct neural network weight parameters developed skills to fill different niches in order to avoid competition within a large population of agents.

As of writing, this project is in active development in the NeurIPS machine learning community~\cite{suarez2021neural} to work towards studies of large agent populations, long time horizons, open-ended tasks, and modular game systems. The developers provide active support and documentation, and also develop additional training, logging, and visualization tools to enable this line of large-scale multi-agent research. This work is still in its early stages, and only time will tell if platforms that enable the study of large populations such as Neural MMO or MAgent gain further traction within the Deep RL communities.

\subsection{Meta-Learning}

In the previous sections, we described works that express the solution to problems in terms of a collection of independent neural network agents acting together to achieve a common goal. These parameters of these neural network models are optimized for the collective performance of the population. While these systems have been shown to be robust and adapt to changes in its environment, they are ultimately hardwired to perform a certain task, and cannot perform another task unless retrained from scratch.

\textit{Meta-learning} is an active area of research within deep learning where the goal is to train the system to learn. It is a large sub-field of ML, including areas such as simple transfer learning from one training set to another. For our purposes, we follow the line of work from Schmidhuber~\cite{schmidhubermeta2020}, where he views meta learning as the problem of ML algorithms that can learn \textit{better} ML algorithms, which he believes is required to build truly self-improving AI systems.

So unlike traditionally training a neural network to perform one task, where the weight parameters of neural networks are traditionally optimized with a gradient descent algorithm, or with evolution strategies~\cite{evojax2022}, the goal of meta-learning is to train a \textit{meta-learner} (which can be another neural network-based system) to learn a \textit{learning algorithm}. This is a particularly challenging task, with a long history {see Schmidhuber~\cite{schmidhubermeta2020} for a review). In this section, we will highlight recent promising works that make use of collective agents that can \textit{learn to learn}, rather than learn to perform only a particular task (which we have covered in the previous section).

Concepts from self-organization can be naturally applied to train neural networks to meta-learn by extending the basic building blocks that compose artificial neural networks. As we know, artificial neural networks consist of identical \textit{neurons} which are modeled as non-linear activation functions. These neurons are connected in a network by \textit{synaposes} which are weight parameters which are normally trained with a learning algorithm such as gradient descent. But one can imagine extending the abstraction of neurons and synapses beyond static activation functions and floating point parameters. Indeed, recent work~\cite{ohsawa2018neuron,ott2020giving} have explored modeling each \textit{neuron} of a neural network as an individual reinforcement learning agent. Using the terminology of RL, each neuron's \textit{observations} are its current state which change as information is transmitted through the network, and each neuron's \textit{actions} enable each neuron to modify its connections with other neurons in the system, hence the problem of learning to learn is treated as a multi-agent RL problem where each agent is part of the collection of neurons in a neural network. While this approach is elegant, the aforementioned works are only capable of learning to solve toy problems and are not yet competitive with existing learning algorithms.

Recent methods have gone beyond using simple scalar weights to transmit scalar signals between neurons. Sandler et al.~\cite{sandler2021meta} introduce a new type of \textit{generalized} artificial neural network where both neurons and synapses have multiple states. Traditional artificial neural networks can be viewed as a special case of their framework with two-states where one is used for activations, the other is used for gradients produced using the backpropagation learning rule. In the general framework, they do not require the backpropagation procedure to compute any gradients, and instead rely on a shared local learning rule for updating the states of the synapses and neurons. This Hebbian-style bi-directional local update rule would only require that each synapse and neuron only requires state information from their neighboring synapse and neurons, similar to cellular automata. The rule is parameterized as a low-dimensional \textit{genome} vector, and is consistent across the system. They employed both evolution strategies, or conventional optimization techniques to meta-learn this genome vector, and their main result is that the update rules meta-learned on the training tasks generalize to unseen novel test tasks. Furthermore, the update rules perform faster than gradient-descent based learning algorithms for several standard classification tasks.

\begin{figure}[ht]
\vskip -0.0in
\begin{center}
\centerline{\includegraphics[width=0.65\columnwidth]{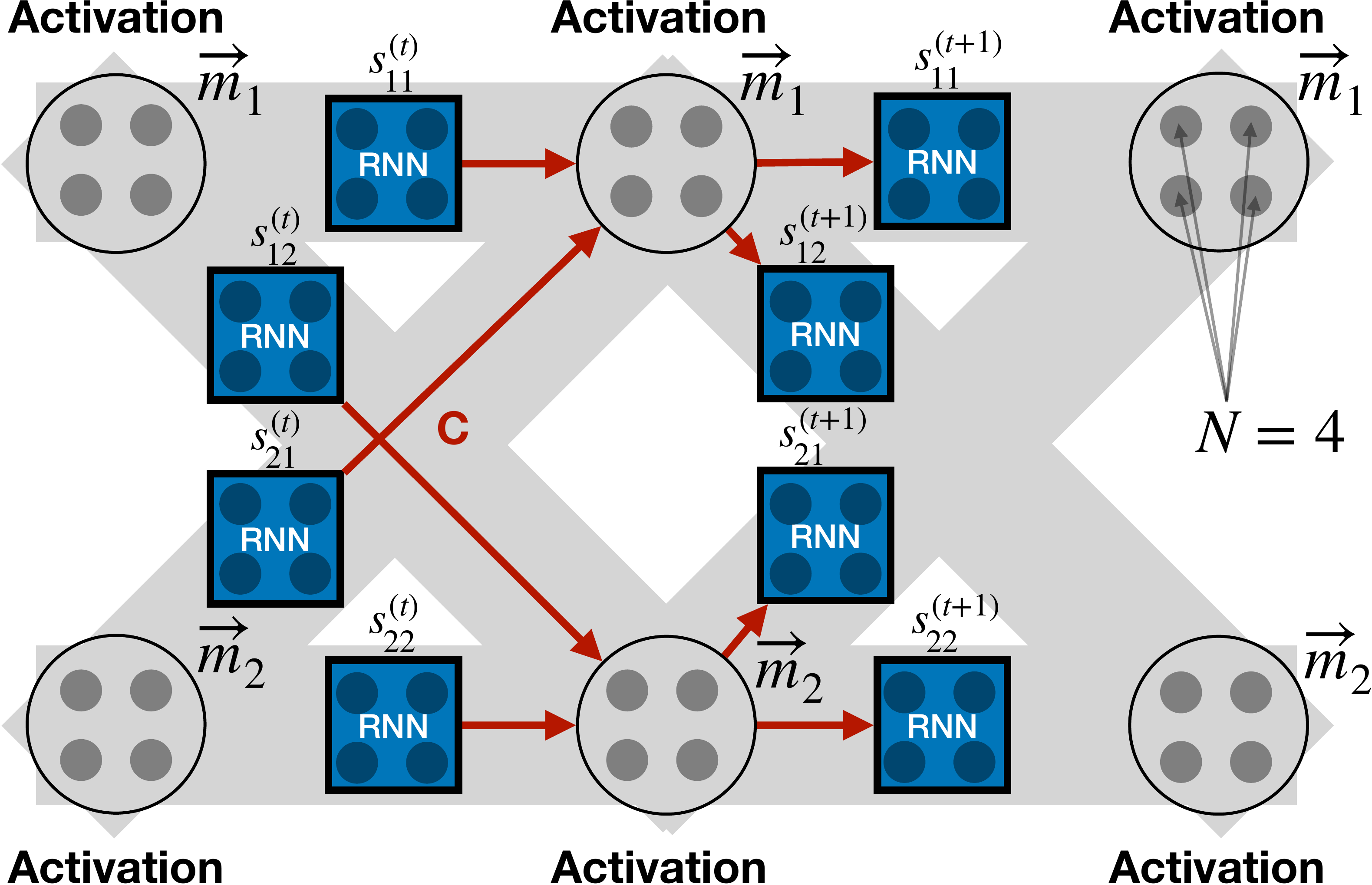}}
\vskip -0.00in
\caption{Recent work by Sandler et al.~\cite{sandler2021meta} and Kirsch et al.~\cite{kirsch2020meta} attempt to generalize the accepted notion of artificial neural networks, where each neuron can hold multiple states rather than a scalar value, and each synapse function bi-directionally to facilitate both learning and inference. In this figure, Kirsch et al.~\cite{kirsch2020meta} use an \textit{identical} recurrent neural network (RNN) (with different internal hidden states) to model each synapse, and show that the network can be trained by simply running the RNNs forward, without using backpropagation.}
\label{fig:kirsch2020meta}
\end{center}
\vskip -0.4in
\end{figure}

A similar direction has been taken by Kirsch et al.~\cite{kirsch2020meta}, where the neurons and synapses of a neural network are also generalized to higher dimension message-passing systems, but in their case each synapse is replaced by an recurrent neural network (RNN) with the same shared parameters. These RNN synapses are bi-directional and govern the flow of information across the network. Like Sandler et al.~\cite{sandler2021meta}, the bi-directional property allows for the network to be used for both inference and learning at the same time by running the system in forward-pass mode. The \textit{weights} of this system are essentially stored in the hidden states of the RNNs so by simply running the system, they can train themselves using the error signals as feedback. Since RNNs are general-purpose computers, they were able to demonstrate that the system can encode the gradient-based backpropagation algorithm by training the system to simply emulate backpropagation, rather than explicitly calculating gradients via hand-engineering. Of course, their system is much more general than backpropagation, and thus capable of learning new learning algorithms that are much more efficient than backpropagation (See Figure~\ref{fig:kirsch2020meta}).

The previous two works mentioned in this section are only recently published at the time of writing, and we believe that these decentralized local meta-learning approaches have the potential to revolutionize the way neural networks are used in the future in a way that challenges the current paradigm that separates model training and model deployment. There is still much work to be done in demonstrating that these approaches can scale to larger datasets, due to inherently much larger memory requirements (due to much larger internal states of the system). Furthermore, while the algorithms are able to produce learning algorithms that are vastly more sample efficient compared to gradient descent, this efficiency is only apparent in the early stages of learning, and performance tends to peak very early on. Gradient descent, while less efficient, is less biased towards few-shot learning, and can continue to run for many more cycles to refine the weight parameters that will ultimately produce networks that achieve higher performance.

\section{Discussion}
\vskip -0.05 in

In this survey, we first gave a brief historical background to describe the intertwined development of deep learning and collective intelligence research. The two research areas were born at roughly the same time, and we can also spot some positive correlations of the rises and falls between the two areas throughout their history. This is no coincidence, since advances and breakthroughs in one of the two areas can usually innovate new ideas or complement the solutions to the problems in the other. For example, introducing deep neural networks and related training algorithms to cellular automata allowed us to develop image generation algorithms that are resistant to noise and have ``self-healing'' properties.
This survey explored several works in deep learning that were also inspired by concepts in collective intelligence.
At a macro-level collective intelligence in multi-agent deep RL led to interesting works that can exceed human performance through collective self-play, and to decentralized self-organizing robot controllers; At a micro-level, collective intelligence is also embedded inside advanced methods of simulating each neuron, synapse or other object at a finer granularity within a system with deep models.

Despite the progress made in the works described in this survey, many challenges lie ahead. While neural CA techniques have been applied to image-processing, their application has so far been limited to relatively small and simple datasets, and their image generation quality is still far below the state-of-the-art on more sophisticated datasets such as ImageNet or Celebrity Faces~\cite{palm2022variational}. For Deep RL, while the surveyed works have demonstrated that a global policy can be replaced by a collection of smaller individual policies, we have yet to transfer these experiments to real physical robots. Finally, we have witnessed self-organization guide meta-learning algorithms. While this line of work is extremely promising, they are currently confined to small-scale experiments due to the large computational requirements that come with replacing every single neural connection with an entire artificial neural network. We believe many challenges will be solved in due time as their trajectories are already in motion.

Looking at their respective development trajectories, DL has accomplished notable achievements in developing novel architectures and training algorithms that led to efficient learning and better performance. The research and development cycle of DL is more engineering-focused, as such the advances seen are more benchmark-based (such as classification accuracy for image recognition problems, or related quantitative metrics for language modeling and machine translation problems). DL advances are generally more incremental and predictable in nature, while CI focuses more on problem formulations and environmental mechanisms that motivate novel emergent group behavior. As we have shown in this survey, CI-based techniques enable new capabilities that were simply not possible before. For instance, it is impossible to \textit{incrementally-improve} a fixed-robot to become a robot capable of self-assembly, and gain all the benefits from such modularity. Naturally, the two areas can complement each other. We are confident that the hand-in-hand style of co-development will continue.


\newpage

\section{Glossary of Terms and Definitions}

\begin{table}[h]
\begin{small}
\begin{tabular}{cl}
\multicolumn{2}{l}{\textit{Deep Learning Related}}                                             

\\

\\ \hline

\textbf{Term} & \textbf{Definition}

\\ \hline
Deep Learning (field)                                                            & \begin{tabular}[c]{@{}l@{}}The study of machine learning methods based on artificial neural\\ networks. Much of the field is devoted to research on the numerous\\ architectures, their training methods, theoretical properties, and\\ applications of artificial neural networks.\end{tabular}                                                                                                                            \\ \hline
Supervised Learning                                                              & \begin{tabular}[c]{@{}l@{}}An approach of learning when both the data and the expected\\ outputs (training signal) are given.\end{tabular}                                                                                                                                                                                                                                                                                  \\ \hline
\begin{tabular}[c]{@{}c@{}}Unsupervised\\ Representation\\ Learning\end{tabular} & \begin{tabular}[c]{@{}l@{}}An approach of learning to represent data in a latent space (the\\ dimension of which is usually, but not necessarily, lower than that\\ of the input data) without additional training signals.\end{tabular}                                                                                                                                                                                    \\ \hline
Transfer Learning                                                                & \begin{tabular}[c]{@{}l@{}}A research problem in ML that focuses on applying knowledge gained\\ from one problem to solve another different but related problem.\end{tabular}                                                                                                                                                                                                                                               \\ \hline
Meta-Learning                                                                    & \begin{tabular}[c]{@{}l@{}}A large sub-field of ML, and in this paper (and including areas such\\ as simple transfer learning from one training set to another). For our\\ purposes, we view meta-learning as the problem of machine learning\\ algorithms that can learn better machine learning algorithms, which\\ many believe is required to build truly self-improving artificial\\ intelligent systems.\end{tabular} \\ \hline
\begin{tabular}[c]{@{}c@{}}Reinforcement\\ Learning\end{tabular}                 & \begin{tabular}[c]{@{}l@{}}RL is an area of ML. It consists of methods that train an agent to\\ improve its policy from interactions with the environment or\\ experiences in order to achieve goals.\end{tabular}                                                                                                                                                                                                          \\ \hline
Agent / Controller                                                               & \begin{tabular}[c]{@{}l@{}}An (artificial) agent or a controller is a system that takes actions\\ corresponding to a series of inputs in order to achieve goals.\end{tabular}                                                                                                                                                                                                                                               \\ \hline
Policy                                                                           & \begin{tabular}[c]{@{}l@{}}A (control) policy is the “guide book” by which an agent makes\\ decisions for its actions. In deep RL, a policy usually takes the\\ form of an artificial neural network which accepts the inputs from\\ the task/environment and outputs the corresponding actions.\end{tabular}                                                                                                               \\ \hline
Self-Play                                                                        & \begin{tabular}[c]{@{}l@{}}A training scheme in RL where an agent is trained by playing\\ against/with snapshots of itself.\end{tabular}                                                                                                                                                                                                                                                                                    \\ \hline
\begin{tabular}[c]{@{}c@{}}Convolutional\\ Neural Networks\end{tabular}          & \begin{tabular}[c]{@{}l@{}}A class of artificial neural networks that are commonly applied to\\ imagery data. Their connectivity pattern resembles the organization\\ of the animal visual cortex.\end{tabular}                                                                                                                                                                                                             \\ \hline
\begin{tabular}[c]{@{}c@{}}Recurrent\\ Neural Networks\end{tabular}              & \begin{tabular}[c]{@{}l@{}}A class of artificial neural networks most commonly applied to analyze\\ sequential/temporal data. RNNs can use their internal states to process\\ inputs of variable lengths.\end{tabular}                                                                                                                                                                                                      \\ \hline
\begin{tabular}[c]{@{}c@{}}Graph\\ Neural Networks\end{tabular}                  & \begin{tabular}[c]{@{}l@{}}A class of artificial neural networks for processing data best represented\\ by graph data structures. Such data examples include social networks,\\ molecule structures, robot morphologies, etc.\end{tabular}                                                                                                                                                                                  \\ \hline
\begin{tabular}[c]{@{}c@{}}Graph\\ Processing\\ Unit\end{tabular}                & \begin{tabular}[c]{@{}l@{}}A GPU is a specialized electronic circuit designed to rapidly accelerate\\ the creation of images. Their highly parallel structure makes them\\ efficient for algorithms that process large blocks of data parallelly and\\ are therefore widely adopted in DL research.\end{tabular}                                                                                                            \\ \hline
MNIST                                                                            & \begin{tabular}[c]{@{}l@{}}MNIST is a dataset of handwritten digits that is commonly used for\\ training image processing systems.\end{tabular}                                                                                                                                                                                                                                                                             \\ \hline
\end{tabular}
\end{small}
\end{table}

\begin{table}[h!]
\begin{small}
\begin{tabular}{ll}
\multicolumn{2}{l}{\textit{Collective Intelligence Related Concepts}}
\\
\\ \hline

\textbf{Term} & \textbf{Definition}

\\ \hline
\begin{tabular}[c]{@{}c@{}}Collective\\ Intelligence (field)\end{tabular}                                                  & \begin{tabular}[c]{@{}l@{}}The study of the shared, or group intelligence that emerges from the\\ interaction (collaboration, collective efforts, and/or competition) of\\ a large group of individuals.\end{tabular}                                                                                                                                                                                                       \\ \hline
Self-Organization                                                                & \begin{tabular}[c]{@{}l@{}}A process where some form of overall order arises from (local)\\ interactions between parts within a system.\end{tabular}                                                                                                                                                                                                                                                                        \\ \hline
\end{tabular}
\end{small}
\end{table}

\begin{table}[h!]
\begin{small}
\begin{tabular}{ll}

\multicolumn{2}{l}{\textit{Other Concepts}}\\
\\ \hline

\textbf{Term} & \textbf{Definition}

\\ \hline
Complex Systems                                                                  & \begin{tabular}[c]{@{}l@{}}Systems whose behavior is intrinsically difficult to model due to the\\ dependencies and interactions between the parts within the system\\ and/or across time.\end{tabular}                                                                                                                                                                                                                     \\ \hline
Cellular Automaton                                                               & \begin{tabular}[c]{@{}l@{}}A CA is a collection of cells on a grid that evolves their states over a\\ set of discrete values according to predefined rules based on the states\\ of the neighboring cells.\end{tabular}                                                                                                                                                                                                     \\ \hline
Embodied Cognition                                                               & \begin{tabular}[c]{@{}l@{}}It is a theory stating that cognition is shaped by aspects of the entire\\ body of the organism. It emphasizes the role of the body (e.g., motor,\\ perception) in forming cognition features (e.g., form concepts, make\\ judgements).\end{tabular}                                                                                                                                             \\ \hline
\end{tabular}
\end{small}
\end{table}

\bibliographystyle{SageH}
\bibliography{main.bib}

\end{document}